\documentclass[10pt,twocolumn,letterpaper]{article}

\usepackage{cvpr}
\usepackage{times}
\usepackage{epsfig}
\usepackage{graphicx}
\usepackage{amsmath}
\usepackage{amssymb}

\usepackage[pagebackref=true,breaklinks=true,letterpaper=true,colorlinks,bookmarks=false]{hyperref}

\cvprfinalcopy % *** Uncomment this line for the final submission

\begin{document}

%%%%%%%%% TITLE
\title{InAR:Inverse Augmented Reality}

\author{Hao Hu\\
Department of Electronics\\
Shan-Dong University\\
Jinan,  China\\
{\tt\small huhao678@126.com}
% For a paper whose authors are all at the same institution,
% omit the following lines up until the closing ``}''.
% Additional authors and addresses can be added with ``\and'',
% just like the second author.
% To save space, use either the email address or home page, not both
\and
Hainan Cui\\
Institute of Automation\\
Chinese Academy of Sciences\\
Beijing, China\\
{\tt\small hncui@nlpr.ia.ac.cn}
}

\maketitle
%\thispagestyle{empty}

%%%%%%%%% ABSTRACT
\begin{abstract}
   Augmented reality is the art to seamlessly fuse virtual objects into real ones. In this short note, we address the opposite problem, the inverse augmented reality, that is, given a perfectly augmented reality scene where human is unable to distinguish real objects from virtual ones, how the machine could help do the job. We show by structure from motion (SFM), a simple 3D reconstruction technique from images in computer vision, the real and virtual objects can be easily separated in the reconstructed 3D scene.
\end{abstract}

%%%%%%%%% BODY TEXT
\section{Introduction}

Augmented reality (AR) is meant to seamlessly fuse interested virtual objects into the real scene, and one of its ultimate goals is to make the fusion imperceptible even by an experienced human being, and great efforts and progresses have been made on it in AR community in recent years
~\cite{Survey}.

In this work, we address the opposite problem: suppose one emerges in a perfectly fused AR environment where he is unable to distinguish which one is virtual, which one is real around him by his own visual system, how could he resort to a machine, for example, his mobile phone or a common digital camera, to aid him to ascertain virtual from real ones? To our knowledge, this inverse problem seems unaddressed in the literature. Here we show by a SFM-based simple 3D reconstruction technique from images captured by his mobile phone camera, the job can be easily done.

\begin{figure*}
\centerline{\includegraphics[width=\textwidth]{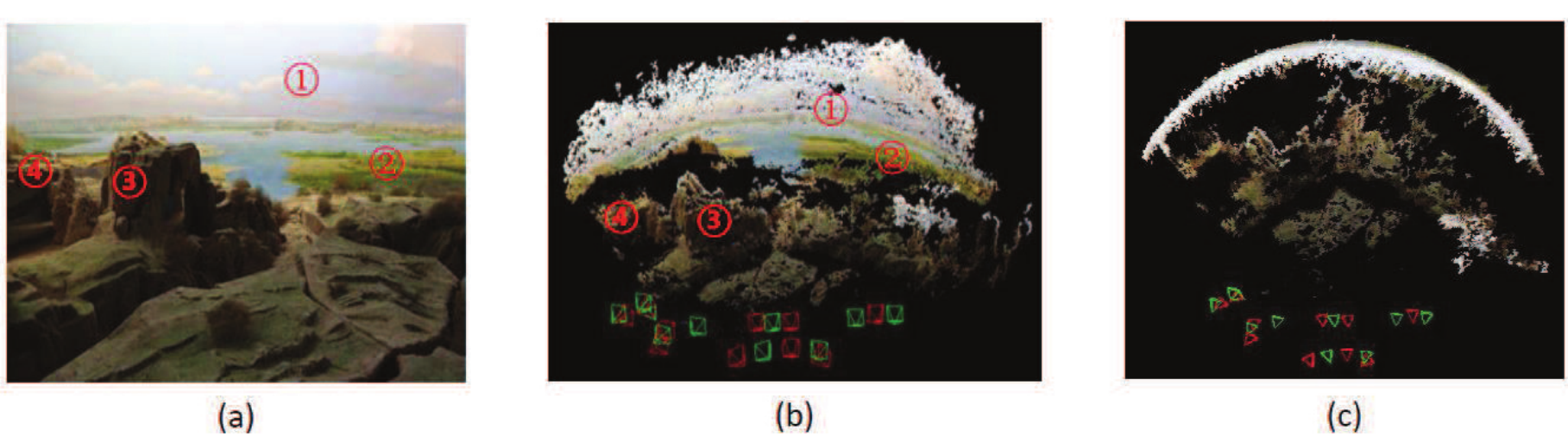}}
\caption{(a): an image of Shui-Dong-Gou Archeological Museum; (b): the dense reconstruction of Shui-Dong-Gou Archeological Museum produced by Bundler+PMVS2; (c): the top-view of the dense reconstruction scene in (b). Note that labels \textcircled{1}、\textcircled{2}、\textcircled{3}、\textcircled{4} in (a) mark some areas in the scene, and their corresponding reconstruction results are respectively showed in (b).}
\label{fig:1}
\end{figure*}

\section{Inverse Augmented Reality}
Shape-from-Motion (SFM) is a widely used technique in 3D scene reconstruction from images in computer vision field. By establishing point correspondences across several images and by some assumptions on the camera~'s intrinsic parameters, currently a dense metric scene reconstruction can be automatically obtained, for example, by Snavely~'s Bundler~\cite{Snavely_Bundler} for sparse reconstruction, followed by Furukawa~'s PMVS~\cite{Furukawa_PMVS} for dense or quasi-dense reconstruction.

In the following, we report a case to illustrate the basic principle and effects. Shui-Dong-Gou Archeological Museum is located in the Ning-Xia-Hui Autonomous Region, North West China, it is an archeological site of human life and activities dating back to about 40,000 years. An example image of this museum is showed in Fig.~\ref{fig:1}(a). Currently its demo hall is in such an exploit of AR technology that ordinary visitors can hardly tell which part of scene is virtual, which part is real.

As shown in Fig.~\ref{fig:1}(a), we can hardly distinguish virtual parts from real scene, at least for the majority of the lay visitors.
By taking about 20 images with a Canon 5D Mark III camera, and by Bundler + PMVS2 combination, the AR scene is 3D reconstructed as shown in Fig.~\ref{fig:1}(b) and Fig.~\ref{fig:1}(c). Note that in Fig.~\ref{fig:1}(b) and Fig.~\ref{fig:1}(c), small colored cones are calibrated camera poses of the used images for our 3D reconstruction, and the Fig.~\ref{fig:1}(c) is the top-view of dense reconstruction.
From Fig.~\ref{fig:1}(c), we can see that the sky marked by \textcircled{1} in Fig.~\ref{fig:1}(a) is in fact a vaulted backdrop, and the greensward marked by \textcircled{2} in Fig.~\ref{fig:1}(a) is also in this backdrop. Stones marked by \textcircled{3}、\textcircled{4} are real ones. In sum, virtual parts can be clearly separated from the real scene from the reconstructed 3D scene.

\section{Conclusion}
Before ending this short note, we would make the following two points:
(1)	In this work, we only discuss a static AR scene, the principle and techniques can be extended straightforwardly to dynamic AR scene using a synchronized stereo system. For example, nowadays many mobile phones are equipped with a micro-array camera, which could capture several images at the same time. From such synchronized images, the same 3D reconstruction technique for the static scene can be used for dynamic scenes.
(2)	Human 3D visual perception is not necessarily truly 3D in the physical sense although the machine and human use similar mechanism for disparity computation ~\cite{Stereoscopic}.

{\small
\bibliographystyle{ieee}
\bibliography{egbib}
}

\end{document}